\documentclass[letterpaper, 10 pt, conference]{ieeeconf}

\IEEEoverridecommandlockouts
\usepackage{graphicx,dblfloatfix}
\usepackage{epsfig}
\usepackage{mathptmx}
\usepackage{times}
\usepackage{amsmath}
\usepackage{amssymb}
\usepackage{flafter}
\usepackage[cmintegrals]{newtxmath}
\usepackage{bm}
\usepackage{pgf}
\usepackage{caption}
\usepackage{subcaption}
\usepackage{algorithm}
\usepackage{algpseudocode}
\usepackage{booktabs}
\usepackage{hyperref}

\DeclareMathOperator*{\argmin}{arg\,min}
\let\amssymbboxplus\boxplus
\let\amssymbboxminus\boxminus
\renewcommand{\boxplus}{\mathbin{\mathop\amssymbboxplus}}
\renewcommand{\boxminus}{\mathbin{\mathop\amssymbboxminus}}

\newcommand\inputpgf[2]{{
\let\pgfimageWithoutPath\pgfimage
\renewcommand{\pgfimage}[2][]{\pgfimageWithoutPath[##1]{#1/##2}}
\input{#1/#2}
}}

\newdimen\figrasterwd
\figrasterwd\columnwidth

\title{\LARGE \bf Multi-Modal Motion Planning\\ Using Composite Pose Graph Optimization}

\author{Lukas Lao Beyer*,
	Nadya~Balabanska*,
	Ezra~Tal,
	and~Sertac~Karaman
	\thanks{*L. Lao Beyer and N. Balabanska contributed equally to this work.}%
	\thanks{All authors are with the Laboratory for Information and Decision Systems, Massachusetts Institute of Technology (MIT), Cambridge, MA 02139, USA.	
		{\tt\small $\{$llb,nadyab,eatal,sertac$\}$@mit.edu}}
}

\begin{document}

\maketitle
\thispagestyle{empty}
\pagestyle{empty}

\begin{abstract}
	
  In this paper, we present a motion planning framework for multi-modal vehicle
  dynamics.  Our proposed algorithm employs transcription of the optimization
  objective function, vehicle dynamics, and state and control constraints into
  sparse factor graphs, which---combined with mode transition
  constraints---constitute a composite pose graph.  By formulating the
  multi-modal motion planning problem in composite pose graph form, we enable
  utilization of efficient techniques for optimization on sparse graphs, such as
  those widely applied in dual estimation problems, e.g., simultaneous
  localization and mapping (SLAM).  The resulting motion planning algorithm
  optimizes the multi-modal trajectory, including the location of mode
  transitions, and is guided by the pose graph optimization process to eliminate
  unnecessary transitions, enabling efficient discovery of optimized mode
  sequences from rough initial guesses.  We demonstrate multi-modal trajectory
  optimization in both simulation and real-world experiments for vehicles with
  various dynamics models, such as an airplane with taxi and flight modes, and a
  vertical take-off and landing (VTOL) fixed-wing aircraft that transitions
  between hover and horizontal flight modes.

\end{abstract}

\section*{SUPPLEMENTARY MATERIAL}
A video of the experiments is available at \url{https://youtu.be/XdUtyGu3p1o}.

\section{INTRODUCTION}

\begin{figure}[t!]
  \centering
  \includegraphics[width=0.98\linewidth]{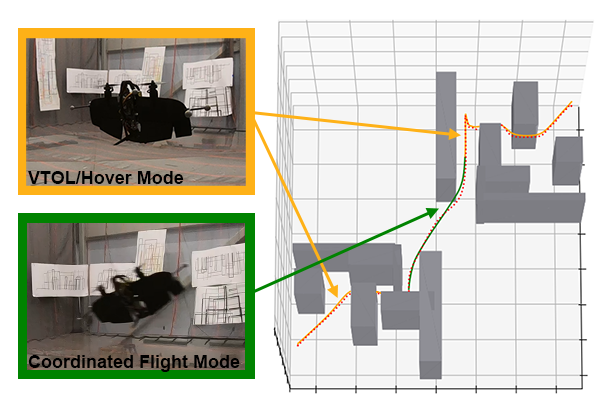}
  \caption{Multi-modal trajectory for a hybrid aircraft, which switches between
    hover mode and coordinated flight mode. The solid and dotted lines indicate
    the planned and real-life flight trajectories, respectively.}
  \label{fig:tailsitter-maze-withpics}
\end{figure}

In this paper, we consider the motion planning problem for robotic vehicles with
multi-modal dynamics. These vehicles switch between multiple modes that each
exhibit dynamics governed by their own respective set of ordinary differential
equations (ODEs).
Mode transitions may occur due to deliberate (mechanical) configuration changes,
e.g. in tiltrotor aircraft that switch between a vertical take-off and landing
(VTOL) configuration and a more aerodynamically efficient cruise
configuration~\cite{tiltrotor}. Future urban mobility concepts may include
vehicles that reduce travel time by combining driving and flying
modes~\cite{urbanmobility}.

Motion planning for multi-modal dynamics has been extensively addressed using
probabilistic sampling-based algorithms, such as
\textit{Multi-Modal-PRM}~\cite{mm-prm} or \textit{Random-MMP}~\cite{random-mmp}
which introduce mechanisms for sampling transitions between modes. However,
having been developed with a focus on robotic manipulation problems involving
contact, these planners do not address challenges posed by dynamics constraints.
Sampling-based approaches like \textit{Sparse-RRT}~\cite{sparse-rrt} can provide
asymptotic optimality guarantees in planning for dynamics-constrained vehicles,
but their reliance on randomly sampled control inputs leads to long planning
times before solutions of acceptable quality are found even for low-dimensional
single-mode systems.
Therefore, we focus on optimization-based motion planning algorithms. These
planners are generally only able to arrive at locally rather than globally
optimal solutions, but algorithms such as \textit{CHOMP}~\cite{chomp},
\textit{STOMP}~\cite{stomp}, or \textit{GPMP}~\cite{gpmp} demonstrate an ability
to rapidly arrive at high-quality solution paths. Timed Elastic Bands
(TEBs)~\cite{teb} perform trajectory optimization by representing trajectories
using a factor graph. This enables the use of efficient graph optimization
techniques, such as those developed for simultaneous localization and mapping
applications~\cite{factor-graphs-in-perception}. Some TEB-based planners are
indeed fast enough to function as a model predictive controller
(MPC)~\cite{teb_mpc}.

Trajectory optimization approaches for multi-modal or hybrid systems often
utilize a hierarchical architecture, combining local trajectory optimization
with a higher-level planner to search over discrete actions. In order to be
computationally tractable, such approaches may depend on simplified dynamics
models~\cite{FernandezGonzalez2018} or learned
approximations~\cite{suh2020energy}, which require significant engineering
effort to design. Many resulting planners may be unsuitable for motion in tight,
cluttered environments~\cite{Saranathan2018, Pajarinen2017}. Other approaches
that do consider obstacles often require elaborate initialization schemes and
are unable to converge without a high-quality initial
guess~\cite{chddp_clutter}.

We propose an algorithm for finding a locally time-optimal trajectory for
multi-modal vehicles based on pose graph optimization. Our approach employs TEBs
to obtain an intuitive and flexible transcription of dynamics, control input,
and obstacle constraints in a sparse hypergraph. To enable application of TEBs
to multi-modal motion planning, we introduce a composite pose graph structure
that combines a sequence of pose graphs each corresponding to a single mode, as
shown in Fig.~\ref{fig:graph}. This allows discovery of optimized mode
transition locations in cases where a mode sequence is known a
priori. Additionally, we propose a scheme for generating an initial mode
sequence and a mechanism for removal of unnecessary mode transitions to enable
automatic discovery of optimized mode sequences.

Our work contains several contributions. Firstly, we present an efficient
minimum-time motion planning algorithm for systems with multi-modal dynamics.
Our planning framework optimizes the mode sequence, configuration space
trajectory and control inputs, and incorporates obstacle avoidance. Secondly, we
extend the TEB planning approach to enable planning with multi-modal
dynamics. To achieve this, we introduce a composite pose graph structure that
enables optimization of mode sequences. Finally, we present elaborate results
that verify our algorithm in both simulation and real-world experiments. For the
real-world experiments, we used a hybrid aircraft that can transition between
VTOL mode with rotorcraft dynamics, and cruise mode with fixed-wing flight
dynamics, as shown in Fig. \ref{fig:tailsitter-maze-withpics}.

The remainder of the paper is structured as follows. In Section
\ref{sec:problem_outline}, we describe the multi-modal planning problem and
provide a detailed overview of our algorithm. This is followed by experimental
results in Section \ref{sec:experiments}, and finally conclusions in Section
\ref{sec:conclusions}.

\section{MULTI-MODAL MOTION PLANNING ALGORITHM}
\label{sec:problem_outline}

We consider multi-modal vehicle dynamics described by a system of ordinary
differential equations (ODEs), as follows:
\begin{equation}\label{eq:dyn1}
\dot x(t) = {f}_{M(t)}(x(t), {u}(t)),
\end{equation}
where $x(t)$, ${u}(t)$, and $M(t)$ are respectively the state, control input,
and dynamics mode as functions of time $t$.  The minimum-time motion planning
problem requires finding the fastest feasible trajectory from the start state
${x}_{\text{start}}$ to the goal state ${x}_{\text{goal}}$.  In additional to
dynamics constraints, it may involve state, control input and obstacle
constraints, leading to the following optimization problem:
\begin{equation}
\label{eq:optim_cont}
\begin{aligned}
&\min_{x(t),\;u(t),\;M(t)} \quad t_e & \\
\textrm{s.t.} \quad & {x}(0) = {x}_{\text{start}},\;{x}(t_e) = {x}_\text{goal} &\textrm{with} \; t_e \geq 0\\
& \dot x(t) = {f}_{M(t)}(x(t), {u}(t)) & \forall t \in [0,t_e]\\
&  x(t) \in \mathcal{X}_{M(t)}   & \forall t \in [0,t_e]\\
&  {{u}}(t) \in \mathcal{U}_{M(t)}   & \forall t \in [0,t_e]\\
&  {{x}}(t) \notin \mathcal{X}^{obs}_{M(t)}   & \forall t \in [0,t_e]\\
\end{aligned}
\end{equation}
where $\mathcal{X}_{M(t)}$ and $\mathcal{U}_{M(t)}$ are box-constrained state
and control input sets corresponding to mode $M(t)$, respectively, and
$\mathcal{X}^{obs}_{M(t)}$ is the corresponding set of obstacles in Euclidean
space.

Our proposed algorithm for finding the solution to \eqref{eq:optim_cont} is
based on transforming the optimization problem into a sparse pose graph
structure using the TEB approach.  The pose graph captures the multi-modal state
dynamics \eqref{eq:dyn1} using resizable trajectory segments.  We use a
probabilistic roadmap (PRM) planner to quickly obtain a trajectory used for
initialization of the pose graph.  The initial trajectory must be accompanied by
an initial mode sequence $\Sigma_\text{init}$.  As described in Section
\ref{sec:init}, our proposed algorithm can find an optimal mode sequence through
pruning under specific conditions.  Mode pruning is performed at each iteration
and followed by trajectory optimization, for which we represent the objective
and constraints from \eqref{eq:optim_cont} using penalty functions.  This
enables the use of efficient methods for unconstrained nonlinear least squares
optimization.  An overview of the complete algorithm is given in Algorithm
\ref{alg:mainalg}, and individual steps are detailed in ensuing sections.

\begin{algorithm}
  \caption{Multi-modal trajectory optimization}
  \label{alg:mainalg}
  \begin{algorithmic}[1]
    \Function{Plan}{$x_{\text{start}}, x_{\text{goal}}, \Sigma_{\text{init}}, N_{\text{iterations}}$} 
    \State $P_{\text{init}} \gets \text{ComputeInitialPath}(x_{\text{start}}, x_{\text{goal}})$
    \State $B \gets $InitializeTEBs$(P_{\text{init}}, \Sigma_\text{init})$
    \For{$i \gets 1$ to $N_{\text{iterations}}$}
    \For{$m \gets 1$ to $|B|$}
    \State ResizeTEB$(B^{[m]})$
    \EndFor
    \State PruneModes$(B)$
    \State Optimize$(B)$
    \EndFor
    \State \textbf{return} ExtractTrajectory($B$)
    \EndFunction
  \end{algorithmic}
\end{algorithm}

\subsection{Pose Graph Formulation} \label{sec:planner}

The vehicle state $x(t)$ may contain state variables governed by higher-order
dynamics, e.g., second-order Newtonian dynamics.  Consequently, $x(t)$ contains
integrator states which we will not include in the pose graph.  Hence, we
eliminate integrator states from $x(t)$ and use the reduced pose vector $q(t)$
in the formulation of the pose graph.  We realize that individual elements of
$q(t)$ may be governed by dynamics of varying order, e.g., second-order vehicle
dynamics combined with first-order control input servo dynamics, and will
address this in the graph formulation.

We divide the trajectory $q(t)$ into single-mode segments and discretize these
into individual poses.  Each single-mode pose segment is captured in a pose
graph, and the composite graph representing the entire trajectory $B$ is
constructed by linking these pose graphs.  A single-mode segment is described by
a tuple, as follows:
\begin{equation} 
B^{[m]}    := (Q^{[m]}, U^{[m]}, \Delta t^{[m]}, M^{[m]}),
\end{equation}
where $Q^{[m]}$, $U^{[m]}$ and $\Delta t^{[m]}$ are respectively the pose,
control input and time step sequences, and $M^{[m]}$ is the segment mode.  We
denote the number of pose nodes in $B^{[m]}$ by $N^{[m]}$.

Penalty functions are used to capture the objective function and constraints in
the pose graph, respectively.  This structure enables formulation of the
trajectory optimization problem as a nonlinear least squares program and the
application of corresponding efficient solvers, as described in Section
\ref{sec:optim}.  The resulting unconstrained objective function is given by
\begin{equation}\label{eq:nlsq}
  \begin{aligned}
  F(B) &= \sum_{m = 1}^{|B|} \sum_{i = 1}^{ N^{[m]}}\left(\alpha_\text{o}{e^{[m]}_{\text{o},i}}^\top e^{[m]}_{\text{o},i} + \sum_\text{c}\alpha_\text{c} {e^{[m]}_{\text{c},i}}^\top e^{[m]}_{\text{c},i} \right) \\
  &+ \sum_{m=1}^{|B|-1} \alpha_{\text{t}} {e^{M^{[m]} \rightarrow M^{[m+1]}}_t}^\top e^{M^{[m]} \rightarrow M^{[m+1]}}_t,
  \end{aligned}
\end{equation}
where $\alpha$ are weights corresponding to the objective and constraint penalty
functions ($e^{[m]}_{\text{o},i}$ and $e^{[m]}_{\text{c},i}$), which are defined
in Table \ref{tab:penalty_functions}.  In this table, a system of ODEs governing
$q$ is denoted with $g$, and its order is denoted with $n$.  Similarly, control
inputs $u$ (with corresponding integrator states) are governed by $h$.  As $x$
may contain integrator states, the state constraint $x \in \mathcal{X}$ and
control input constraint $u \in \mathcal{U}$ can technically apply to $q$ and
$u$ and their derivatives up to $q^{(n-1)}$ and $u^{(n-1)}$.  If this is the
case, additional functions $g$ and $h$, each with varying order, may be defined
according to the form given for \textit{Vehicle Dynamics} in Table
\ref{tab:penalty_functions}.  The resulting value of the penalty $e_c$ for the
dynamics is proportional to the difference between the output of the nominal
dynamics given by given by $g$ or $h$, and the configuration derivative (e.g.,
$q^{(n)}$) obtained from the current pose graph.

State derivatives are obtained using the finite difference method, as reflected
by the sequence of connected nodes.  The start and goal states are directly
substituted into constant first and final pose nodes and thus enforced through
the dynamics constraint.  Similarly to the state constraint $x \in \mathcal{X}$,
$x_\text{start}$ and $x_\text{goal}$ may constrain the initial and final
velocities, accelerations, and higher derivatives up to order $n-1$.
Consequently, these boundary constraints may require substitution into vehicle
dynamics penalty terms connected to the first and final $n$ poses.  Obstacles
are defined in the Euclidean space and thus only pertain to $q$ itself, so that
the Euclidean distance to the closest obstacle can be taken as
$\text{ObstacleDistance}$.  Finally, the penalty functions
$e^{M^{[m]} \rightarrow M^{[m+1]}}_t$ impose continuity constraints at the mode
transitions.

\begin{figure}
  \centering
  \includegraphics[width=0.85\linewidth]{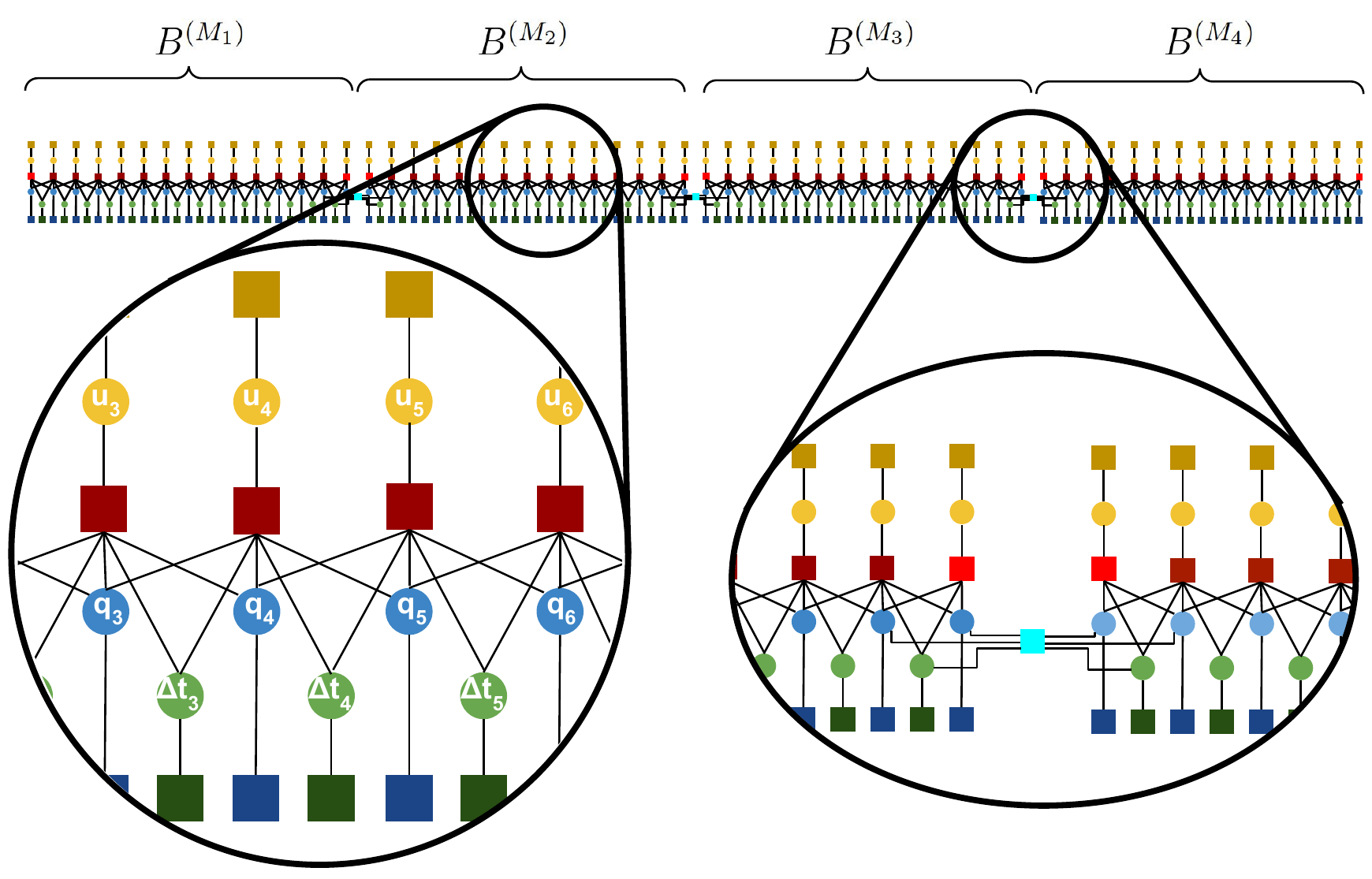}
  \caption{Factor graph representation of the cost of a four-mode trajectory
    $B = [B^{[M_1]}, B^{[M_2]}, B^{[M_3]}, B^{[M_4]}]$. Round nodes represent
    vertices containing optimizable parameters, squares with links represent
    edges of the factor graph and are colored according to the class of their
    corresponding cost function (red: second order dynamics, blue: obstacle
    avoidance, yellow: control input limit, green: time minimization, cyan: mode
    transition).}
  \label{fig:graph}
\end{figure}

\begin{table*}[!ht]
  \centering
  \vspace{1em}
  \caption{Transformation of trajectory optimization objective and constraints to pose graph.}
  \label{tab:penalty_functions}
  \setlength{\tabcolsep}{4pt}
  \begin{tabular}{lllll}
    \hline
    Objective/Constraint & &Penalty Term ($e^{[m]}_{\text{o},i}$ or $e^{[m]}_{\text{c},i}$) & Connected Nodes \\
    \hline
    Time Minimization & $\min t_e$ & $\Delta t_i$& $\Delta t_i$ \\
    Start and Goal States & ${x}(0) = {x}_{\text{start}},\;{x}(t_e) = {x}_\text{goal}$ & Substitution into first and final pose nodes.&&\\
    Vehicle Dynamics & $\dot x(t) = {f}_{M(t)}(x(t), {u}(t))$& \begin{tabular}{@{}l@{}} $| q^{(n)}_i - {g}_{M^{[m]}}(q_i,\dots,q^{(n-1)}_i, {u}_i) |$ \\ $| u^{(n)}_i - {h}_{M^{[m]}}(u_i,\dots,u^{(n-1)}_i) |$ \end{tabular}& \begin{tabular}{@{}l@{}} $q_{i-\lfloor \frac{n}{2} \rfloor},\dots,q_{i+\lceil \frac{n}{2} \rceil}, \Delta t_{i-\lfloor \frac{n}{2} \rfloor},\dots,\Delta t_{i+\lfloor \frac{n}{2} \rfloor}, u_i$ \\ $u_{i-\lfloor \frac{n}{2} \rfloor},\dots,u_{i+\lceil \frac{n}{2} \rceil}, \Delta t_{i-\lfloor \frac{n}{2} \rfloor},\dots,\Delta t_{i+\lfloor \frac{n}{2} \rfloor}$ \end{tabular} \\
    State Feasibility & $x(t) \in \mathcal{X}_{M(t)}$&  $\max(0, q_i - q^{\text{max}}_{M^{[m]}}) + \max(0, q^{\text{min}}_{M^{[m]}} - q_i)$ & $q_i$\\
    Control Input Feasibility & $ {{u}}(t) \in \mathcal{U}_{M(t)}$ &  $\max(0, u_i - u^{\text{max}}_{M^{[m]}}) + \max(0, u^{\text{min}}_{M^{[m]}} - u_i)$ & $u_i$ \\
    Obstacle Avoidance & ${{x}}(t) \notin \mathcal{X}^{obs}_{M(t)}$ & $\max(0, -\text{ObstacleDistance}(M(t),q_i))$& $q_i$\\
    \hline
  \end{tabular}
\end{table*}
$F(B)$ depends only sparsely on the elements of the trajectory $B$.  Represented
as a hypergraph, each $q_i$, $u_i$ and $\Delta T_i$ are taken to be the
vertices, connected by edges corresponding to the various penalty functions.
Fig. \ref{fig:graph} illustrates this graph as well as its sparse, periodic
structure.  Finding the optimal trajectory $B^{\star}$ then amounts to solving
the following unconstrained nonlinear least squares program
\begin{equation}\label{eq:opt}
B^{\star} = \argmin_{B}{F(B)}.
\end{equation}
We found that assigning the weights $\alpha_c$, $\alpha_t$ and $\alpha_o$ such
that the minimum-time objective does not compete with minimization of dynamics
residuals or obstacle avoidance (i.e., $\alpha_t \approx \alpha_c \gg \alpha_o$)
gives favorable results.  With such a weight assignment, the optimization is
initially guided towards satisfying dynamic feasibility and obstacle
constraints.  Once residuals corresponding to these constraints have been
minimized significantly, the time minimization cost (with lower weight) becomes
a significant contributor to the overall cost function, guiding the optimization
towards lower time solutions.

\subsection{Initialization}\label{sec:init}

An initial trajectory with corresponding mode sequence is required to construct
and initialize the factor graph.  We found the algorithm to be generally
insensitive to state and control input feasibility and time-optimality of the
initial trajectory, but somewhat sensitive to satisfaction of the obstacle
avoidance constraint.  Hence, we opt for a sampling-based planner on
$\mathbf{SE}(3)$, specifically PRM*, to quickly obtain a collision-free initial
trajectory~\cite{rrtstar}.  The required mapping from $\mathbf{SE}(3)$ to the
state $q$ is trivial for most vehicle dynamics applications, and linear and
spherical linear interpolation (Slerp) are used to obtain continuous position
and orientation between PRM* waypoints.  Time differences are initialized to a
small positive value, and control inputs are initialized to the mean of their
bounds.  As shown in Section \ref{sec:experiments}, the algorithm obtains
reliable convergence despite the infeasibility of the initial trajectories.

The initial mode sequence may be based on engineering knowledge of the system
and scenario.  Additionally, if the vehicle mode transition graph is fully
connected, our algorithm is able to arrive at any subsequence of the initial
mode sequence through mode pruning.  Hence, it is able to find an optimal
transition sequence after initialization with a sequence that repeatedly loops
through all modes.  Suppose we expect at most $N_{\text{transitions}}$
transitions are required between modes $M_1, \ldots, M_{N_{\text{m}}}$, then
such a looping mode sequence is constructed as
\begin{align*}
\Sigma = [\underbrace{M_1, \dots, M_{N_{\text{m}}}}_{1}, \dots, \underbrace{M_1, \dots, M_{N_{\text{m}}}}_{N_{\text{transitions}}}].
\end{align*}
The initial geometric trajectory is then divided in single-mode segments of
equal length.  After initialization, redundant and unnecessary modes are
incrementally eliminated during the trajectory optimization process, as shown in
Fig. \ref{fig:pruning-merging}.  The complexity of this method scales linearly
with the number of modes and transitions, and thus it enables our algorithm to
avoid the combinatorial explosion from which mode sequencers may suffer. We
verified its favorable scaling in systems with a limited number of modes, but
expect it to be impractical for systems with a very large number of modes, which
require an excessive initial mode sequence.

\subsection{Graph Optimization}\label{sec:optim}

As given by (\ref{eq:opt}), the motion planning process is formulated as an
unconstrained minimization of the cost function $F(B)$.  We chose the
Levenberg-Marquardt (LM) algorithm~\cite{LM} to solve this nonlinear least
squares problem.  For notational simplicity, we will limit the description in
this section to a single-mode trajectory, i.e., $|B|=1$.  There is no
fundamental difference between this and multi-modal cases.

Starting at an initial guess, LM iteratively linearizes around $B$ and finds an
increment $\Delta \mathbf{x}$ to be applied to the trajectory.  This increment
is represented as a vector containing the increments $\Delta q_i$, $\Delta u_i$
and $\Delta \Delta T_i$, which are applied as follows:
\begin{align}
  B \boxplus \Delta \mathbf{x}  &= (Q \boxplus \Delta Q, U \boxplus \Delta U, \tau \boxplus \Delta \tau)
                                  \label{eq:traj-with-increment} \\
  Q \boxplus \Delta Q  &= \{ q_i \boxplus \Delta q_i       \}_{i= 0 \ldots N}   \\
  U \boxplus \Delta U  &= \{ u_i \boxplus \Delta u_i       \}_{i= 0 \ldots N}   \\
  \tau \boxplus \Delta \tau   &= \{ \Delta T_i + \Delta\Delta T_i \}_{i= 0 \ldots N-1} \label{eq:delta-t-increment}
\end{align}
We use the $\boxplus$ operator to denote the element-wise application of the
increments. When using $\boxplus$ to apply increments to $q_i$ or $u_i$, its
exact behavior depends on the representation of the increments, configurations,
and control inputs. For example, for some of the scenarios described in Section
\ref{sec:experiments}, the configurations $q_i$ are taken to be poses in
\textbf{SE}(3) with the rotation stored as a quaternion, but the increments
$\Delta q_i$ contain the rotational increment in a non-overparametrized
representation (the axis of a normalized quaternion), since small increments are
far from the singularities encountered in such a representation. Hence, the
$\boxplus$ operator behaves like motion composition in this case~\cite{boxplus}.

Using a first-order approximation for the cost terms
${e}_{c, i}(B^{[m]} \boxplus \Delta \mathbf{x}) \approx {e}_{c, i}(B^{[m]}) +
{J}_{c, i} \Delta \mathbf{x}$ with ${J}_{c, i} $ the Jacobian of the penalty
function, we obtain approximations of the cost after increment per penalty $c$
and pose index $i$ as
\begin{align}
  F&_{c,i}(B \boxplus \Delta \mathbf{x}) \nonumber \\
     =&~ \alpha_c {e}_{c,i}(B \boxplus \Delta {x})^{\top} {e}_{c,i}(B \boxplus \Delta \mathbf{x})  \\
     \approx&~ \alpha_c ({e}_{c,i}(B) + {J}_{c,i} \Delta \mathbf{x})^{\top} ({e}_{c,i}(B) + {J}_{c,i} \Delta \mathbf{x})
       \nonumber \\
     =&~\alpha_c (
       \underbrace{{e}_{c,i}(B)^{\top} {e}_{c,i}(B)}_{c_{c,i} / \alpha_c}
       + 2 \underbrace{{e}_{c,i}(B)^{\top} {J}_{c,i}}_{{{b}_{c,i}}^\top / \alpha_c} \Delta \mathbf{x}
       + \Delta \mathbf{x}^{\top} \underbrace{{{J}_{c,i}}^{\top} {J}_{c,i}}_{{H}_{c,i} / \alpha_c} \Delta \mathbf{x})\nonumber
       \label{eq:category-error-approx} 
       \end{align}
A similar form is obtained for the objective cost, so that
\begin{align}
  F&(B \boxplus \Delta \mathbf{x}) \nonumber \\
     &\approx \sum_{c,i} c_{c,i} + 2 {{b}_{c,i}}^{\top} \Delta \mathbf{x} + \Delta \mathbf{x}^{\top} {H}_{c,i} \Delta \mathbf{x}\\
     &+\sum_i c_{o, i} + 2 {{b}_{o, i}}^{\top} \Delta \mathbf{x} + \Delta \mathbf{x}^{\top} {H}_{o, i} \Delta \mathbf{x}\nonumber \\
     &= c + 2 {b}^{\top} \Delta \mathbf{x} + \Delta \mathbf{x}^{\top} {H} \Delta \mathbf{x}\nonumber \label{eq:total-error-approx}
\end{align}
Next, the increment $\Delta \mathbf{x}^\star$ is obtained by solving
\begin{align}
  ({H} + \lambda {I}) \Delta \mathbf{x}^\star = -{b},
\end{align}
where $\lambda$ is the damping factor.  At $\lambda = 0$, the Gauss-Newton
algorithm is recovered. However, to better handle nonlinear costs, LM adaptively
sets $\lambda$ to control the maximum size of the increments
$\Delta \mathbf{x}^\star$.

Due to the structure of the cost function and its representation as factor graph
shown in Fig. \ref{fig:graph}, each of the individual Jacobians $J_{c, i}$ only
depends on very few elements of $B$ and $\Delta \mathbf{x}$, leading to sparsity
of ${H}$.  The $\mathrm{g}^2\mathrm{o}$ library~\cite{g2o} was chosen for
implementation of the motion planner, as it allows specification of the cost
function as a graph, and can thus exploit its sparse structure to efficiently
perform the optimization.

\subsection{Incremental Trajectory Resizing} \label{sec:resizing}

The shape and time of the trajectory may significantly change during
optimization, leading to changing position and time differences between
consecutive poses.  Hence, it is necessary to insert and delete poses in order
to maintain efficiency, and accurate finite-differencing and obstacle avoidance.

Position-based removal ensures that resolution is always sufficient for
satisfactory obstacle avoidance. Given distance thresholds $d_{min}$ and
$d_{max}$ and a function $\mathrm{Position}(q_i)$ that maps the pose to a point
in Euclidean space, a pose $q_i$ (with corresponding control input and time
difference) is removed if
$\left|\mathrm{Position}(q_i) - \mathrm{Position}(q_{i+1})\right| < d_{min}$.
The new time difference $\Delta T_i$ (previously $\Delta T_{i+1}$) is
incremented by the removed time difference to maintain identical trajectory
time.

When
$\left|\mathrm{Position}(q_i) - \mathrm{Position}(q_{i+1})\right| > d_{max}$, a
new pose is inserted between $q_i$ and $q_{i+1}$. The newly inserted pose
$q_{i+1}$ and control input $u_{i+1}$ are computed using system-dependent
interpolation functions $q_{i+1} = \mathrm{Interp}_q(q_{i}, q_{i+2}, 0.5)$ and
$u_{i+1} = \mathrm{Interp}_u(u_{i}, u_{i+2}, 0.5)$.  In typical vehicle motion
planning scenarios, $q_i$ represents poses in \textbf{SE}(3), such that
$\mathrm{Interp}_q$ performs linear interpolation for the translational part of
the pose, and Slerp for the rotational component.  For the vector-valued control
inputs $u_i$, $\mathrm{Interp}_u$ performs linear
interpolation. Time-differences are each set to half of the previous time
difference.

An analogous procedure is used for time-difference based adjustments.  Here,
$\Delta T_{min}$ and $\Delta T_{max}$ trigger removal or insertion of poses in
order to maintain a time-resolution adequate for good fidelity of the dynamics
model.

\subsection{Mode pruning}\label{sec:pruning}

\begin{figure}
  \centering
  \includegraphics[width=\linewidth]{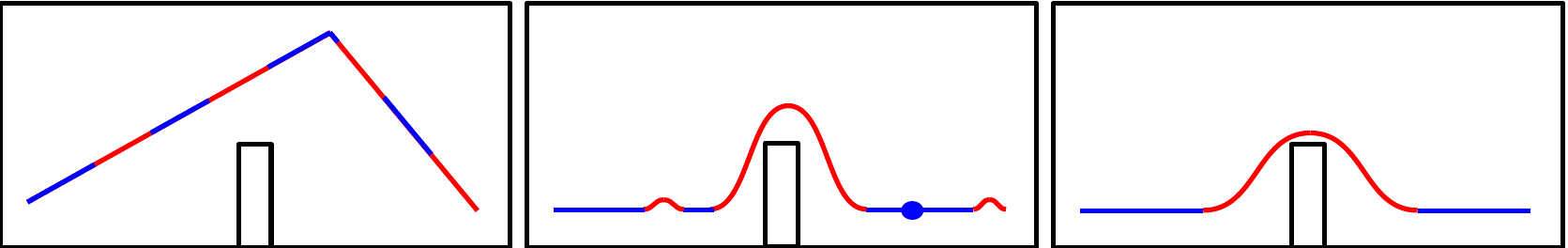}
  \caption{Pruning of modes of which the trajectories were eliminated through
    the resizing process, and merging of trajectories of the same mode that
    become adjacent due to pruning of intermediate modes. The left graphic shows
    an initial guess for a multi-modal trajectory, with colors representing
    separate modes. The middle graphic illustrates the partially optimized
    trajectory in which the third trajectory of the red mode has been resized
    away. The right graphic shows the fully optimized trajectory after pruning
    and merging.}
  \label{fig:pruning-merging}
\end{figure}

The resizing process described in Section \ref{sec:resizing} often leads to
expansion or shrinkage of the single-mode trajectories that compose the full
multi-modal trajectory. This process can result in trajectories being
iteratively shrunk until only their first and last poses remain. To fully
eliminate such a trajectory segment $B^{[n]}$, our algorithm prunes them from
the factor graph by deleting any remaining poses and adding the relevant
transition constraints between the adjacent trajectories $B^{[n-1]}$ and
$B^{[n+1]}$. If the trajectories $B^{[n-1]}$ and $B^{[n+1]}$ correspond to the
same mode, they are combined into a single trajectory instead of being joined by
a transition constraint. This process allows discovery of an optimized mode
sequence in cases where the initialization contains unnecessary mode switches,
as illustrated in Fig. \ref{fig:pruning-merging}.

\section{EXPERIMENTAL RESULTS}
\label{sec:experiments}

We now construct motion planners for a variety of single-mode and multi-modal
systems to experimentally demonstrate our approach. All experiments were
executed single-threaded on an Intel Core i7 2600k CPU, and Jacobians were
computed numerically.  A video of the experiments is available at
\url{https://youtu.be/XdUtyGu3p1o}.

\subsection{Dubins Car}
To illustrate the effectiveness of our dynamics transcription approach, we first
consider a simple single-mode Dubins car.  Graph-based planning for a Dubins car
is also demonstrated by \cite{teb}, but their proposed method requires
hand-designed constraints based on path geometry. Our systematic transcription
of the dynamics results in a different pose graph formulation. In particular, we
consider a car model with bounded control inputs for acceleration $a_\text{des}$
and steering angle $\alpha$.
Fig. \ref{fig:car-rrtstar-comparison} shows planning results and a comparison of
our proposed algorithm with RRT*~\cite{rrtstar}. The RRT* steering function is
based on Dubins geometry with turning radius equivalent to that of the model
used in our planner at the saturated steering angle.
\begin{figure}[htb]
  \centering
   \parbox{\figrasterwd}{
    \parbox{.4\figrasterwd}{%
      \subcaptionbox{\label{fig:car-rrtstar-comparison:a} Optimized trajectory using FGO approach alongside RRT* solutions.}{%
        \resizebox{\hsize}{!}{\input{figures/car-rrt.pgf}}
        \vspace*{-1em}}%
    }
    \hskip0.5em
    \parbox{.5\figrasterwd}{%
      \vspace*{-1em}
      \subcaptionbox{Control inputs $\alpha_i$ and $a_{\text{des}, i}$ along the FGO solution trajectory.\label{fig:car-rrtstar-comparison:b}}{%
        \resizebox{1.075\hsize}{!}{\input{figures/car-traj-u.pgf}}
        \vspace*{-1em}}%
      \vskip1em%
      \subcaptionbox{\label{fig:car-rrtstar-comparison:c} Runtime comparison}{
        \setlength{\tabcolsep}{3pt}
        \small
        \begin{tabular}{lll}
          \toprule
          Planner & Time [s] & Path length \\
          \midrule
          \color{red} FGO & 3.5 & 51.176 \\
          \color{orange} RRT* & 6 & 56.705 \\
          \color{green} RRT* & 500 & 53.406 \\
          \color{blue} RRT* & 1500 & 52.577 \\
          \bottomrule
        \end{tabular}
        }}
    }
  
    \caption{Trajectory planning for the Dubins car vehicle using our factor
      graph optimization based planner (FGO) and
      RRT*. Fig. \ref{fig:car-rrtstar-comparison:a} shows result trajectories
      color coded corresponding to the entries in
      Fig. \ref{fig:car-rrtstar-comparison:c}, which lists planning times for
      each result (FGO result includes PRM-based initialization, which requires
      $<0.1$ seconds).  Note that the FGO optimization recovers control inputs
      which resemble the optimal bang-bang policy
      (Fig. \ref{fig:car-rrtstar-comparison:b}).  }
  \label{fig:car-rrtstar-comparison}
\end{figure}

When starting from an low-quality, potentially dynamically infeasible, initial
trajectory, our proposed planner may take several seconds to converge, as shown
in Fig.~\ref{fig:car-rrtstar-comparison:c}.  We observed that this convergence
time is strongly reduced if a high-quality initial trajectory is available, such
as when performing replanning to incorporate changed start and goal states, or
small variations in the obstacle map.  The capability to efficiently utilize
previous solutions, gives our algorithm potential for online applications where
the planned trajectory must be continuously updated.

\subsection{Airplane} \label{sec:airplane}

We plan trajectories for an airplane which can both fly and taxi on the
ground. This multi-modal system is described by the same dynamics as the Dubins
car for the \textit{taxi} mode, and uses a point-mass aircraft model in the
\textit{flight} mode. The flight mode represents the pose of the airplane using
$(x, y, z)$ for the position, pitch angle $\theta$ and heading angle $\psi$. The
control input is given by the thrust, relative lift coefficient and a roll
angle. The flight mode dynamics include a second-order lift and drag force model
\cite{airplane-model}. When transitioning from taxi to flight, a mode transition
constraint ensures continuity in the velocity as well as pitch and yaw angle.

\begin{figure}[htb]
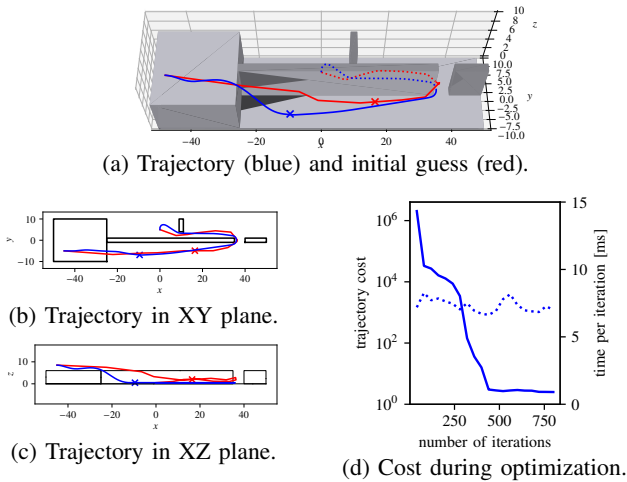

  \centering
  \parbox{\figrasterwd}{
    \parbox{\figrasterwd}{%
      \subcaptionbox{Trajectory (blue) and initial guess (red).\label{fig:carplane-3d}}{%
        \resizebox{\hsize}{!}{\input{figures/carplane-3d.pgf}}
        \vspace*{-1em}}%
    }
    \parbox{\figrasterwd}{%
      \parbox{0.47\figrasterwd}{%
        \vskip0.1em%
        \subcaptionbox{Trajectory in XY plane.\label{fig:carplane-xy}}{%
          \resizebox{\hsize}{!}{\input{figures/carplane-xy.pgf}}
          \vspace*{-0.75em}}%
        \vskip0.1em
        \subcaptionbox{Trajectory in XZ plane.\label{fig:carplane-xz}}{%
          \resizebox{\hsize}{!}{\input{figures/carplane-xz.pgf}}
          \vspace*{-0.75em}}%
      }
      \hskip1em
      \parbox{0.47\figrasterwd}{%
        \subcaptionbox{Cost during optimization.\label{fig:carplane-opt}}{%
          \resizebox{\hsize}{!}{\input{figures/carplane-opt.pgf}}
          \vspace*{-1em}}%
      }
    }
  }
  
  \caption{Trajectory optimization for a taxiing and flying airplane. Transition
    from taxi mode to flight mode is indicated by cross markers.}
  \label{fig:carplane-traj}
\end{figure}

Fig. \ref{fig:carplane-traj} shows an optimized trajectory for this multi-modal
system. In the example, the taxi mode is preferred, such that the vehicle stays
on the ground for as long as possible, only taking off to fly over an obstacle
and reach the goal which is not accessible in taxi mode.  This is achieved by
replacing the original time-optimality cost with an energy consumption cost.
Energy consumption is modelled by assigning a fixed power usage to each mode,
with the power required by the flight mode set to be significantly higher than
that of the taxi mode.  Since this modification is equivalent to assigning the
time-optimality cost for the flight mode to be higher than that of the taxi mode
by a constant multiplier, this energy consumption cost is straightforward to
model within our framework.

\subsection{Tailsitter}

\begin{figure}[htb]
  \centering
  \parbox{\figrasterwd}{
    \parbox{1.0\figrasterwd}{%
      \centering
      \subcaptionbox{Planned (blue) and flown (red) tailsitter trajectories. \label{fig:tailsitter-zigzag}}{%
        \makebox[\hsize][c]{\resizebox{0.6\hsize}{!}{\input{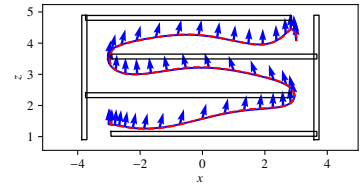}}}}%
    }%
    \vskip1em
    \parbox{1.0\figrasterwd}{%
      \centering
      \subcaptionbox{Planned control inputs along the trajectory.\label{fig:tailsitter-zigzag-controls}}{%
        \makebox[\hsize][c]{\resizebox{0.5\hsize}{!}{\input{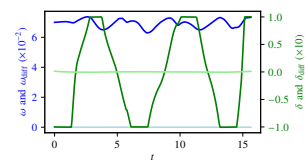}}}}%
    }
  }

  \caption{Optimized trajectory of 329 poses for tailsitter using $\phi$-theory
    model. Convergence time was 10 seconds.}
  \label{fig:tailsitter-zigzag-traj}
\end{figure}

To examine the planner's performance in optimizing trajectories for more
realistic vehicle dynamics, we implement the \textit{$\phi$-theory} model
presented in \cite{Lustosa2019}. This model describes the dynamics of a
tailsitter VTOL aircraft including post-stall behavior using polynomial
equations. It provides 6 degree-of-freedom equations of motion as a second order
ODE, with four control inputs $\omega_L$, $\omega_R$, $\delta_L$ and $\delta_R$
which correspond to left and right propeller angular velocities, and left and
right elevon deflections, respectively. Fig. \ref{fig:tailsitter-zigzag-traj}
shows an example trajectory planned for this model, as well as results from
flying the trajectory with the tailsitter aircraft described in
\cite{bronz2020}.  In the flight experiments, we used the planned position,
velocity, acceleration, attitude and angular rate as reference for the on-board
flight controller.  The consistency between planned and real-life trajectories
demonstrates the applicability of our planning algorithm to complicated models
that realistically capture complex dynamics using an increased number of state
variables.

\subsection{Multi-Modal Tailsitter}

As demonstrated above, our planner is able use the $\phi$-theory tailsitter
model to create high quality trajectories incorporating transitions between
hover and level flight.  However, using such a global description of the vehicle
dynamics may lead to optimized paths that are undesirable due to restrictions in
the control scheme on the vehicle.  For example, planned paths might contain
sideways ``knife-edge'' flight which is difficult for a controller to steadily
track.  Therefore, it is useful to consider a planner which treats the
tailsitter as a multi-modal vehicle capable of transitioning between a
multicopter-like hover and a coordinated flight modes.
Fig. \ref{fig:tailsitter-maze-withpics} shows planning and real-life flight
results for multi-modal trajectories planned using the airplane model from
Section \ref{sec:airplane} to describe coordinated flight, and a simple
multicopter model with a two-dimensional control input consisting of pitch and
roll angles to describe hover.

\section{CONCLUSION}
\label{sec:conclusions}

Our algorithm addresses the motion planning problem for multi-modal vehicle
dynamics using a pose graph formulation. Leveraging efficient pose graph
optimization algorithms, it is able to produce trajectories for a wide variety
of vehicles. We verify its performance and applicability through extensive
computational experiments and real-life experiments using a VTOL aircraft. While
the algorithm does not guarantee discovery of an optimal mode sequence without
an appropriate initialization, it does demonstrate the ability to rapidly
approach optimal trajectories and mode sequences in a variety of realistic
scenarios.

\bibliographystyle{IEEEtran}
\bibliography{./main.bib}

\end{document}